\documentclass[lettersize,journal]{IEEEtran}
\usepackage{amsmath,amsfonts}
\usepackage{array}
\usepackage[caption=false,font=normalsize,labelfont=sf,textfont=sf]{subfig}
\usepackage{textcomp}
\usepackage{stfloats}
\usepackage{url}
\usepackage{verbatim}
\usepackage{cite}
\usepackage{enumitem}
\usepackage{amsthm}
\usepackage{amsmath}
\usepackage{graphicx}
\usepackage{multirow}
\usepackage{tabularx}
\usepackage{pifont}
\usepackage{color}
\usepackage{balance}

\usepackage{booktabs}
\usepackage[table]{xcolor}
\usepackage{float} 
\usepackage[ruled,vlined,linesnumbered]{algorithm2e}
\definecolor{mygreen}{RGB}{243,231,219}
\definecolor{myblue}{RGB}{199,209,213}
\begin{document}

\title{Seeing Time: Benchmarking Chronological Reasoning and Shortcut Biases in Vision-Language Models}

\author{Haoyu Zhou,
Qing Qing,
Caichong Li,
Qixin Zhang,~\IEEEmembership{Graduate Student Member,~IEEE,}
Yongcheng Jing,
Ziqi Xu,~\IEEEmembership{Member,~IEEE,}
Juncheng Hu,
Xikun Zhang,
Renqiang Luo,~\IEEEmembership{Member,~IEEE,}
\thanks{Haoyu Zhou, Qing Qing, Caichong Li, Juncheng Hu and Renqiang Luo are with College of Computer Science and Technology, Jilin University, Changchun 130012, China (\{zhouhy2223, qingqing25, licc1123\}@mails.jlu.edu.cn, \{jchu, lrenqiang\}@jlu.edu.cn).}
\thanks{Qixin Zhang is with College of Computing and Data Science, Nanyang Technological University, 639798, Singapore (e-mail: qixin.zhang2026@gmail.com).}
\thanks{Yongcheng Jing is with School of Computer Science, Wuhan University, China (e-mail:ycjing@zju.edu.cn)}
\thanks{Ziqi Xu, Xikun Zhang, and Feng Xia are with the School of Computing Technologies, RMIT University, Melbourne, VIC 3000, Australia (e-mail: \{ziqi.xu, xikun.zhang\}@rmit.edu.au, f.xia@ieee.org).}

\thanks{Corresponding author: Xikun Zhang, Renqiang Luo.}}


\maketitle

\begin{abstract}
Recent advancements in Vision-Language Models (VLMs) have significantly enhanced their ability to interpret complex visual semantics, yet their capacity for chronological reasoning remains under-explored. In this paper, we introduce a novel benchmark specifically designed to evaluate how VLMs perceive and reason about chronological information within and across images. Unlike existing video-based benchmarks that focus on frame sequencing, our work delves into the underlying logic of chronological judgment and the expansion toward multimodal integration. To facilitate this, we construct three specialized datasets: one containing visually similar objects spanning long historical durations, another categorized by diverse event and object types, and a third pairing images with time-sensitive news text for cross-modal alignment. Through extensive experiments, we analyze whether models exhibit performance disparities across categories and, crucially, explore whether they rely on ``incorrect shortcuts'', such as image color rather than genuine chronological features. Our results reveal that while VLMs show promise, they frequently exploit superficial cues like grayscale versus color filters to bypass authentic chronological reasoning. By providing these high-quality datasets and a rigorous evaluation framework, we offer a diagnostic tool to identify current limitations and guide the development of more robust, logically grounded multimodal models. The source code is shown in~\url{https://github.com/LuoRenqiang/ChronoVision}.
\end{abstract}


\section{Introduction}
\par The rapid advancement of Large Language Models has fundamentally transformed the landscape of visual information processing, ushering in the era of powerful Vision-Language Models (VLMs)~\cite{jian2024large}. 
Unlike traditional computer vision techniques that often rely on narrow, task-specific labels, modern VLMs demonstrate a remarkable ability to interpret complex semantic relationships within images~\cite{antol2015vqa, rasekh2025enhancing}. 
These models now play a pivotal role in bridging the gap between raw pixels and high-level human reasoning across a multitude of industries~\cite{huang2023language}. 
Consequently, the capacity of VLMs to perceive and analyze nuanced information has become a cornerstone of current artificial intelligence research~\cite{hu2024bliva, zhang2025mme}.

\par As these models move toward deployment in sensitive real-world applications, it is essential to scrutinize their internal logic and decision-making evidence~\cite{celona2025improving}. 
Understanding how a model arrives at a specific judgment, especially when dealing with multidimensional information like time, is a prerequisite for building trust and reliability~\cite{imam2025can}. 
We must determine if these models can extend their reasoning beyond simple recognition to handle complex, multimodal temporal cues~\cite{tang2025lego, luo2026bridging}. 
Evaluating this capability is a vital step in ensuring that VLMs can function effectively in dynamic environments where chronological context is key~\cite{cai2025comparebench}.

\par Despite the proliferation of general benchmarks, existing evaluations of temporal reasoning in VLMs remain largely insufficient and narrow in scope~\cite{chu2024timebench}. 
Most current tests are confined to sequencing frames within a single video clip, which fails to explore the deeper reasoning foundations or cross-modal expansion of the models~\cite{zhao2024set, cheng2025caparena}. 
Specifically, it remains unclear whether the genuine chronological reasoning dictates model output or if the models rely on incorrect shortcuts, such as mistakenly associating black-and-white photography with older eras regardless of the actual content. 
Furthermore, there is a significant lack of research into whether models can truly synchronize visual evidence with chronological information found in unstructured text.

\par To address these gaps, we have constructed three specialized datasets to systematically probe the limits of chronological reasoning in VLMs. 
The first dataset contains a vast library of images featuring similar objects that span significant historical periods to challenge the model’s fine-grained recognition of design evolution. 
The second dataset categorizes images by distinct event and object types to facilitate a rigorous analysis of performance consistency across different domains. 
Finally, the third dataset integrates contemporary news data corresponding to specific events to evaluate the model’s ability to align textual chronological markers with visual content.

\par
\begin{figure} [t]
    \centering
    \includegraphics[width=0.40\textwidth]{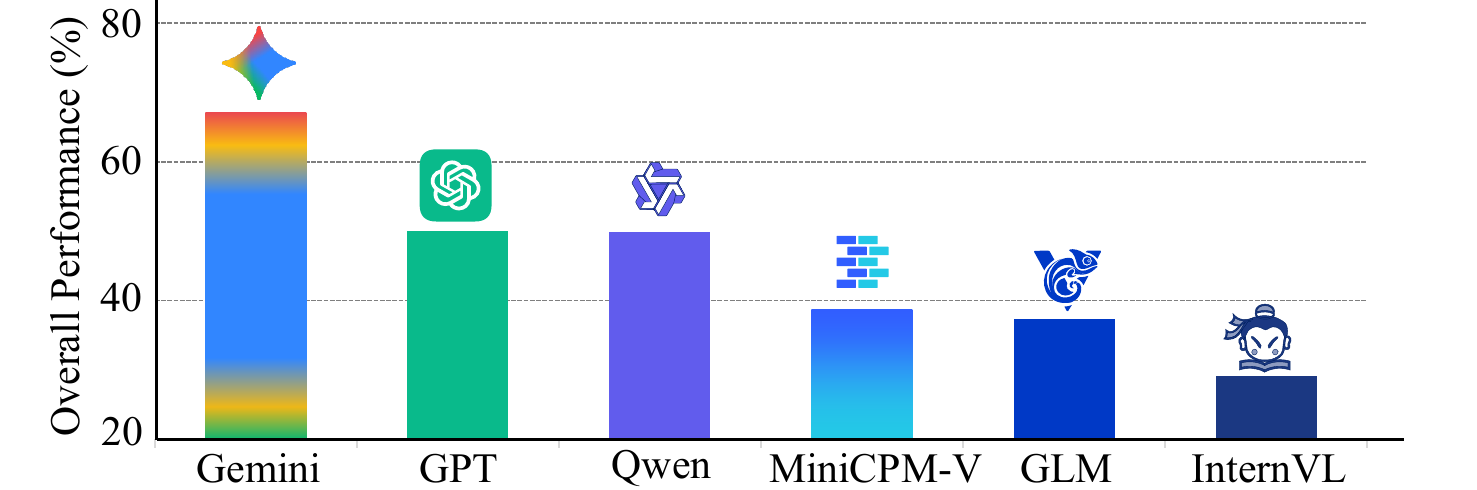}
    \caption{Overall performance of six VLMs across the proposed benchmark. 
    Scores are integrated from the Artifacts, Shortcut, and News tasks.}
    \label{fig:placeholder}
    \vspace{-1em}
\end{figure}

 Utilizing these datasets, we conducted extensive experiments centered on three primary research dimensions (as shown in Fig.~\ref{fig:placeholder}). 
Our analysis first investigates whether VLMs exhibit significant performance disparities when inferring time across different categories of objects or events. 
We then delve into whether the models rely on spurious visual shortcuts, such as image color, to bypass genuine chronological reasoning. 
Lastly, we assess the capability of these models to accurately align and fuse chronological information across both textual and visual modalities.
In summary, our contributions are as follows:
\begin{itemize} [leftmargin=0.5cm]
    \item We propose a novel evaluation framework that shifts the focus from simple video frame sequencing to a deeper exploration of chronological reasoning and multimodal expansion in VLMs. 
    This framework establishes a standard for assessing whether models truly understand chronological evidence or merely perform surface-level recognition.
    \item We construct three datasets featuring long-span object evolution, categorized event types, and news-aligned image-text pairs. 
    These datasets are engineered to expose performance disparities across domains and test the model's ability to synchronize visual cues with textual metadata.
    \item Our extensive empirical analysis identifies critical ``incorrect shortcuts'', such as the model's over-reliance on image color rather than genuine chronological features. 
    These insights quantify current limitations in multimodal alignment and provide a clear roadmap for developing more authentic reasoning capabilities in future models.
\end{itemize}

\section{Related Work}
\subsection{Chronology Reasoning}
\par Chronological reasoning, the ability to understand and sequence events or objects over time, is a fundamental aspect of human intelligence and has recently attracted increasing attention in LLM research~\cite{yan2026investigating, feng2024far}. 
Existing studies mainly focus on temporal knowledge memorization and internal reasoning mechanisms. 
For example, ChroKnowBench~\cite{parkchro2025knowledge} evaluates chronologically accumulated knowledge across different temporal states and shows that LLMs often fail to correctly recall knowledge evolution over time. 
Similarly, MedChangeQA~\cite{vladika2025facts} reveals that LLMs frequently rely on outdated medical consensus when handling rapidly evolving healthcare knowledge. 
Mechanistic studies further identify Temporal Heads~\cite{park2025does}, specialized attention circuits responsible for temporal knowledge processing. 

\par Despite these advances, chronological reasoning in the visual domain remains under-explored. 
Existing video-based benchmarks mainly focus on frame sequencing and fail to capture the deeper logic of chronological judgment across static images. 
To address this limitation, we introduce a benchmark specifically designed to evaluate chronological reasoning abilities in VLMs.

\subsection{VLMs Evaluation Benchmarks} 
\par With the rapid development of VLMs, numerous benchmarks have been proposed to evaluate multimodal perception and reasoning abilities~\cite{liu2021visual, tekaya2025matter}. 
Early benchmarks mainly focused on perception tasks such as image captioning and object recognition, while recent works emphasize more complex reasoning capabilities~\cite{marino2019ok}. 
For example, MMBench~\cite{liu2024mmbench} introduces a bilingual evaluation pipeline with CircularEval to provide stable and objective assessments. 
To improve evaluation reliability, MMStar~\cite{chen2024we} constructs visually indispensable datasets that reduce reliance on language priors, while SimpleVQA~\cite{cheng2025simplevqa} evaluates factuality across diverse scenarios. 
In addition, CAPTURE~\cite{dong2024benchmarking} improves detailed image captioning evaluation through expert annotations and fine-grained metrics.

\par Recent benchmarks further extend to multi-image understanding and temporal relations~\cite{yue2024mmmu}. 
For instance, MuirBench~\cite{wang2025muirbench} evaluates diverse multi-image relations, including temporal ordering, through pairwise unanswerable variants. 
However, existing benchmarks mainly focus on perception or simple frame relations, often overlooking knowledge-intensive chronological reasoning and long-term object evolution~\cite{sun2023journeydb}. 

\par Our work addresses this gap by introducing a benchmark for chronological reasoning in VLMs. 
Unlike prior video-based evaluations that mainly test frame ordering, our benchmark emphasizes long-span historical evolution and cross-modal news alignment. 
Moreover, we design a diagnostic framework to identify whether models rely on superficial shortcuts, such as color, rather than genuine chronological understanding.

\section{Data Construction and Processing}
\par
\begin{figure*}[t]
    \centering
    \includegraphics[width=0.80\textwidth]{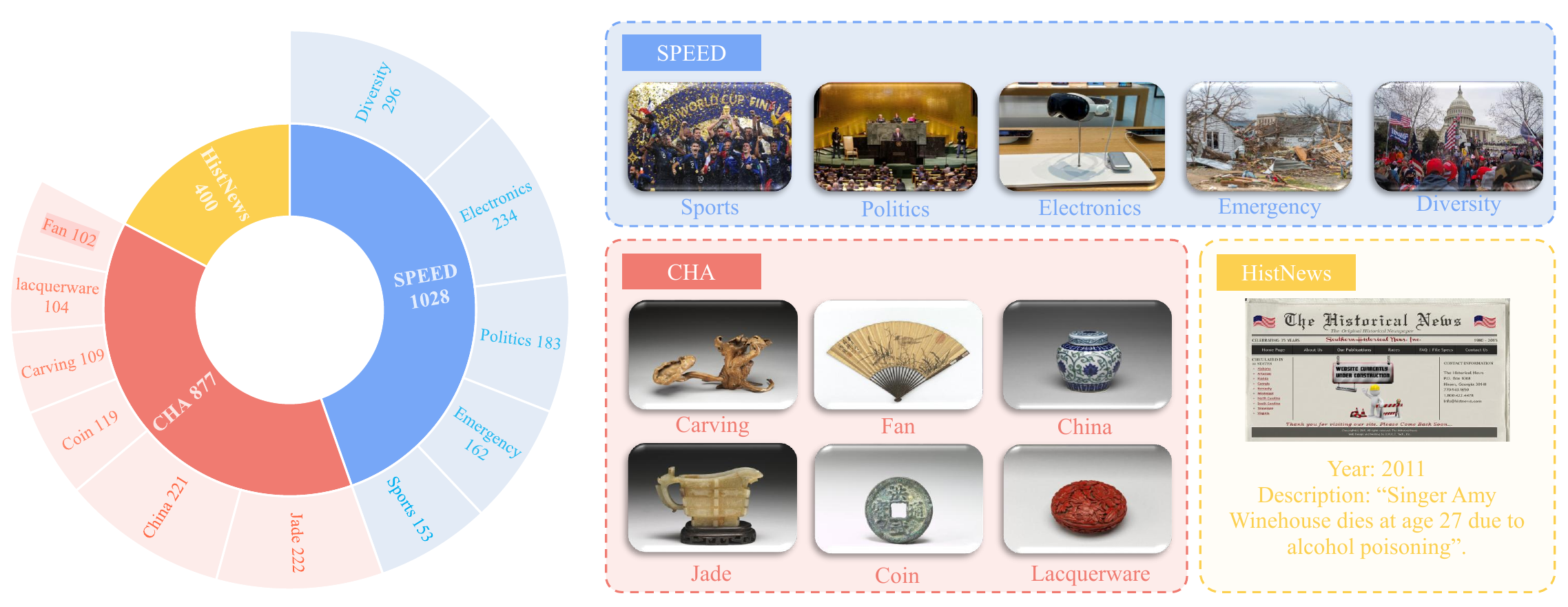}
    \caption{Sample images and data structures from the proposed benchmark, including the Artifacts (CHA), Shortcut (SHEEP), and News tasks (HistNews).}
    \label{fig:dataset}
    \vspace{-1em}
\end{figure*}

 In this section, we provide a description of the data collection and processing pipelines used to construct our multi-faceted benchmark for chronological reasoning.
Example instances from each dataset are illustrated in Fig.~\ref{fig:dataset}.
We meticulously curated three distinct sub-datasets to ensure a rigorous evaluation of how VLMs handle chronological cues across various contexts and modalities. 
The construction process involved large-scale data harvesting, automated cleaning to remove low-quality samples, and manual verification to ensure high chronological fidelity and categorical accuracy. 
By integrating long-span object evolutions, diverse event types, and paired news-text metadata, we establish a robust foundation for diagnosing model performance and identifying reliance on visual shortcuts. 
This systematic approach ensures that our test provides a challenging and diverse environment for assessing state-of-the-art VLMs in multimodal chronological reasoning.

\subsection{The Chinese Historical Artifacts (CHA) Dataset}
The CHA dataset is a specialized collection designed for the Artifacts Task, which serves as the cornerstone for evaluating deep historical chronology reasoning. 
As illustrated in Table~\ref{tab:data_cha}, this dataset comprises $887$ high-resolution images meticulously selected to span five pivotal Chinese dynasties: the Tang, Song, Yuan, Ming, and Qing, capturing the varying prevalence of specific crafts throughout history.
To challenge the model's fine-grained recognition of design evolution, the data is categorized into six distinct types of artifacts, each reflecting the gradual aesthetic and technological shifts of its respective era. 
These categories include Fans, characterized by flat surfaces often adorned with landscape paintings; Jade, representing the symbolic ``beautiful stones'' of Chinese culture; and China, consisting of porcelain fired at high temperatures with vitreous glazes. 
Additionally, the dataset includes Coins, defined by their iconic square-holed round geometry; Lacquerware, featuring intricate protective raw lacquer coatings; and Carvings, which represent the visual artistry of material removal using wood.

\par The integrity of the CHA dataset is rooted in its rigorous sourcing and authentication process, which ensures that subtle stylistic transitions are preserved for model testing. 
All data were harvested from official government open platforms\footnote{https://data.gov.tw/}, providing a reliable foundation of primary historical evidence.
Crucially, to meet the high standards of academic and professional accuracy required for ancient chronological studies, every image and its associated metadata underwent a secondary round of verification and annotation by museum experts\footnote{https://www.npm.gov.tw/}.
This expert involvement ensures that the dataset accurately captures the chronological markers, such as the evolution of porcelain glaze patterns or the shifting geometry of jade carvings, that distinguish one dynasty from the next. 
By providing a clean, expert-verified corpus, the CHA dataset poses a significant challenge for VLMs to move beyond superficial recognition and engage in authentic, fine-grained reasoning across centuries of cultural development.
\begin{table}[t]
  \centering
  \footnotesize
  \caption{Composition of the CHA dataset.}
  \label{tab:data_cha}
  \begin{tabular}{lrrrrrr}
    \toprule
    Category & Tang & Song & Yuan & Ming & Qing & Total \\
    \midrule
    carving & $11$ & $10$ & $3$ & $35$ & $50$ & $109$ \\
    Jade & $22$ & $50$ & $50$ & $50$ & $50$ & $222$ \\
    china & $21$ & $50$ & $50$ & $50$ & $50$ & $221$ \\
    coin & $50$ & $50$ & $0$ & $3$ & $16$ & $119$ \\
    fan & $0$ & $0$ & $12$ & $50$ & $50$ & $112$ \\
    lacquerware & $0$ & $4$ & $0$ & $50$ & $50$ & $104$ \\
    \midrule
    Total & $104$ & $164$ & $115$ & $238$ & $266$ & $887$ \\
    \bottomrule
  \end{tabular}
    \vspace{-1em}
\end{table}

\par To further validate the quality of the proposed datasets and examine whether they contain a meaningful learnable signal beyond memorization, we additionally establish a supervised fine-tuning (SFT) baseline using the open-source model Qwen$2.5$-VL-$7$B.
Specifically, we fine-tune the model on the training split and evaluate it on the held-out test set. 
As shown in Table~\ref{tab:sft_results}, the fine-tuned model consistently outperforms the zero-shot baseline across nearly all dynasties, improving the average accuracy from $43.43$\% to $56.57$\%. 
Overall, these results confirm that the benchmark contains a substantial learning signal and is suitable for evaluating temporal reasoning and chronological understanding in VLMs.

\begin{table}[t]
    \centering
    \caption{Performance comparison between zero-shot inference and supervised fine-tuning (SFT) using Qwen2.5-VL-7B on CHA.}
    \label{tab:sft_results}
    \footnotesize
    \begin{tabular}{lcccccc}
        \toprule
        Method & Tang & Song & Yuan & Ming & Qing & Avg \\
        \midrule
        Zero-Shot & $47.62$ & $57.58$ & $0.00$ & $42.22$ & $52.83$ & $43.43$ \\
        SFT-Test & $57.14$ & $66.67$ & $8.70$ & $60.00$ & $67.92$ & $56.57$ \\
        \bottomrule
    \end{tabular}
    
    \vspace{-1em}
\end{table}

\subsection{The Sports, Politics, Electronics, Emergency, and Diversity (SPEED) Dataset}
\par The SPEED dataset is a meticulously curated chronological collection designed to facilitate a rigorous analysis of performance consistency across different domains. 
Each data instance in this dataset is composed of a high-quality color image containing implicit chronological cues, an associated ground-truth timestamp, and a concise textual title summarizing the visual content. 
To ensure both authenticity and modern relevance, we utilized an automated crawler to collect $2,077$ camera-captured photos from Wikimedia Commons\footnote{https://commons.wikimedia.org}, spanning a broad temporal range from $1952$ to $2025$. 
This extensive timeframe allows the benchmark to evaluate a model's sensitivity to evolving visual styles and technological advancements over more than seven decades.

\par To maintain the integrity of the chronology reasoning task, we manually filtered the dataset to retain $1,028$ high-quality images. 
During this process, we purposefully excluded non-photographic content such as maps and charts, as well as ambiguous samples like generic cityscapes that lack sufficient indirect cues for chronological localization. 
Most importantly, we removed any images containing explicit chronological leaks, such as visible year digits on billboards or signs, to ensure the model relies on genuine visual reasoning rather than shortcuts based on Optical Character Recognition (OCR). 
This rigorous selection process yielded a retention rate of $49.49\%$, ensuring that SPEED serves as a challenging and robust environment for evaluating chronological localization capabilities.

\par
\begin{figure}[t]
    \centering
    \includegraphics[width=0.30\textwidth]{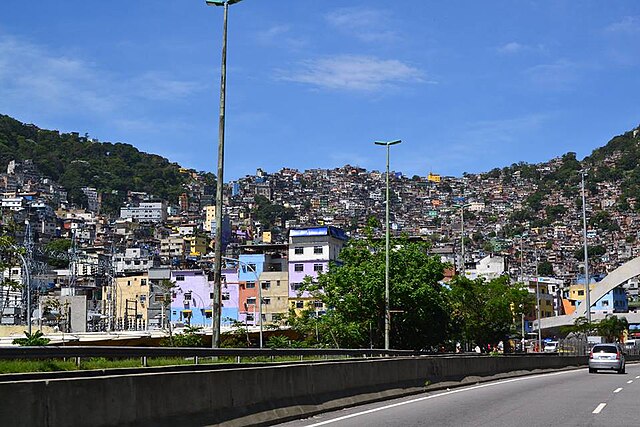}
    \caption{A photo of cityscapes excluded from the Politics.}
    \label{fig:speed_landscape}
    \vspace{-1em}
\end{figure}

 Fig.~\ref{fig:speed_landscape} serves as a representative example of a generic cityscape that was excluded during our manual filtering process to ensure the robustness of the SPEED benchmark. 
While the image provides a clear view of urban density and architectural styles, it lacks sufficient indirect cues necessary for precise chronological localization. 
Specifically:
\begin{itemize}  [leftmargin=0.5cm]
    \item Chronological Ambiguity: The scene depicts a densely populated favela that has maintained a consistent appearance over several decades, making it difficult to distinguish between the late $20$th century and the present day.
    \item Absence of Diagnostic Markers: There are no visible diagnostic artifacts, such as specific vehicle models, contemporary technological infrastructure, or unique fashion styles, that would allow a model to perform rigorous chronological reasoning.
    \item Inherent Noise: The inclusion of such samples would likely encourage models to rely on spurious guesses or statistical priors rather than legitimate visual deduction.
\end{itemize}

\par Consequently, we removed this and similar ambiguous cityscapes, ensuring that every image in the final set contains verifiable, albeit indirect, chronological evidence.

\par
\begin{figure}[t]
    \centering
    \includegraphics[width=0.30\textwidth]{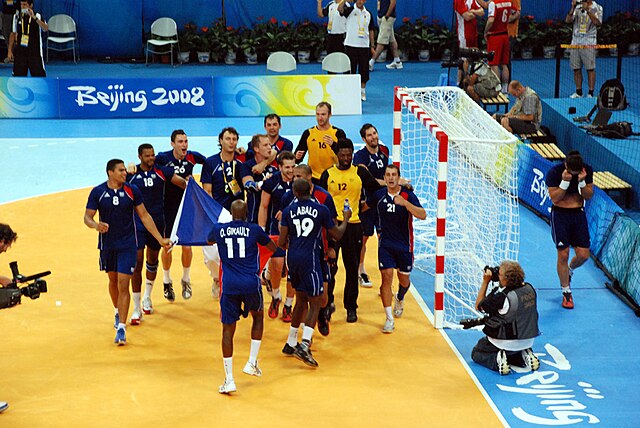}
    \caption{A photo with OCR excluded from the Sports.}
    \label{fig:speed_ocr}
    \vspace{-1em}
\end{figure}

 We also strictly excluded images containing explicit chronological leaks that would allow a model to bypass reasoning through simple OCR.
As shown in Fig.~\ref{fig:speed_ocr}, the provided example:
\begin{itemize} [leftmargin=0.5cm]
    \item Direct Year Labeling: The image contains clear, large-scale text explicitly stating ``Beijing $2008$''.
    \item Elimination of OCR Shortcuts: Such visible year digits on banners, billboards, or signs act as a "shortcut" for multimodal models. Including them would evaluate a model's OCR capabilities rather than its ability to infer chronology from visual context.
    \item Task Integrity: By removing these instances, we ensure the model must rely on authentic chronological reasoning (such as analyzing the jerseys, the event atmosphere, or the technological era of the camera equipment), rather than simply reading the answer from the background.
\end{itemize}

\par The dataset is organized into five distinct subsets that represent various facets of human history and modern life. 
The Sports and Politics categories capture major global milestones, ranging from the FIFA World Cup to APEC summits, while the Electronics subset features iconic consumer products, including recent releases like the iPhone $17$, to test contemporary chronological awareness.
The Emergency subset documents significant natural disasters, such as the Wenchuan earthquake, providing high-stakes visual contexts for time-stamping. 
Finally, the Diversity subset includes a broad collection of varied, timestamp-verified images that further challenge the model’s ability to generalize its chronological reasoning across heterogeneous visual domains.

\subsection{HistNews}
To further investigate the integration of multi-modal information, we present HistNews, a curated textual dataset comprising $400$ significant historical events spanning the period from $1946$ to $2025$. 
This dataset is designed to evaluate a model's ability to align contemporary news data and textual chronological markers with corresponding visual content. 
For each year within this eight-decade range, we meticulously selected five representative events from Wikipedia, ensuring they are evenly distributed across different months to maintain a high degree of chronological diversity and granularity. 
Every entry in the collection underwent rigorous manual verification to guarantee factual accuracy and historical relevance, providing a reliable ground truth for cross-modal testing.

\par The data is structured in a standardized JSON format, with each instance containing four key attributes: the specific year, the full date, a detailed description of the event, and the source URL for verification. 
Within our broader evaluation framework, HistNews serves as the foundational data for the News Task, a critical subtask that challenges VLMs to bridge the gap between linguistic time indicators and visual evidence. 
By requiring the model to map descriptive historical narratives to their correct chronological windows, this dataset probes whether VLMs possess a true understanding of how global events unfold over time. 
Ultimately, HistNews provides the necessary context to determine if models can go beyond simple image recognition and achieve sophisticated, text-driven chronological alignment.

\begin{figure}[t]
    \centering
    \includegraphics[width=0.40\textwidth]{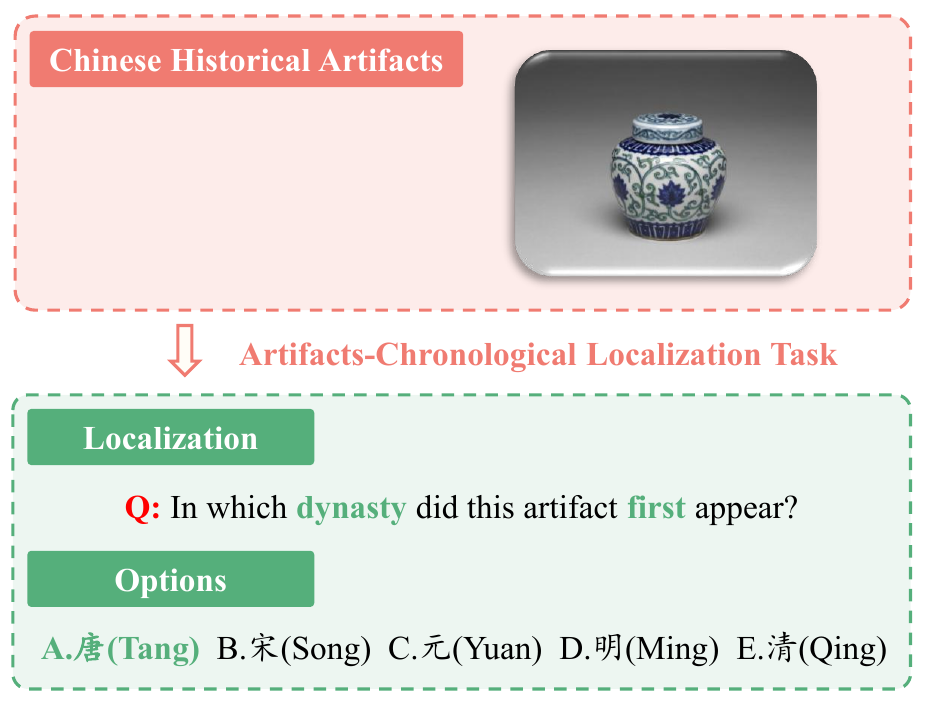}
    \caption{An Example of the Artifacts-Chronological Localization Task.}
    \label{fig:alt}
    \vspace{-1em}
\end{figure}

\section{Methodology}
\par In this section, we introduce a series of benchmark tests designed to comprehensively evaluate the chronological reasoning, chronological localization, and multimodal comprehension capabilities of current VLMs. 
Our evaluation framework is structured around three distinct subtasks that directly map to our core research questions: the Artifacts Task, the Shortcut Task, and the News Task. 
To eliminate any potential interference from external metadata or file-naming conventions, we implemented a strict randomization protocol where all image files were renamed with arbitrary numerical identifiers.
Furthermore, we ensured that the ground-truth answers were randomly and evenly distributed across all tasks to prevent models from exploiting frequency biases or positional patterns. 
This rigorous experimental setup guarantees that the resulting performance metrics reflect the models' genuine ability to process chronological information rather than their capacity for superficial pattern matching.

\subsection{Artifacts Task}
\par The Artifacts Task is designed to evaluate the proficiency of VLMs in chronological reasoning and dynasty estimation, specifically focusing on the evolutionary trajectory of ancient Chinese artifacts. 
As established in the CHA dataset, the evaluation space is constrained to five dynasties: the Tang, Song, Yuan, Ming, and Qing. 
This task investigates the performance consistency of models across different historical eras and material categories through two distinct experimental setups.

\par The first subtask, Artifacts–Chronological Localization, assesses the model's ability to identify the specific historical period of an individual artifact. 
Formally, let $\mathcal{A}$ denote the image collection from the CHA dataset and $P$ represent a standardized prompt inquiring about the dynasty of origin. 
Given an image $I \in \mathcal{A}$, the model must select the correct dynasty from five candidate options in a multi-choice format. 
This setup serves as a direct measure of the model's discriminative power in recognizing subtle stylistic and material features, such as glaze patterns or carving techniques, that are indicative of specific dynastic transitions.
An illustrative sample of the task interface and the localization requirements is provided in Fig. ~\ref{fig:alt}.

\par
\begin{figure}[t]
    \centering
    \includegraphics[width=0.40\textwidth]{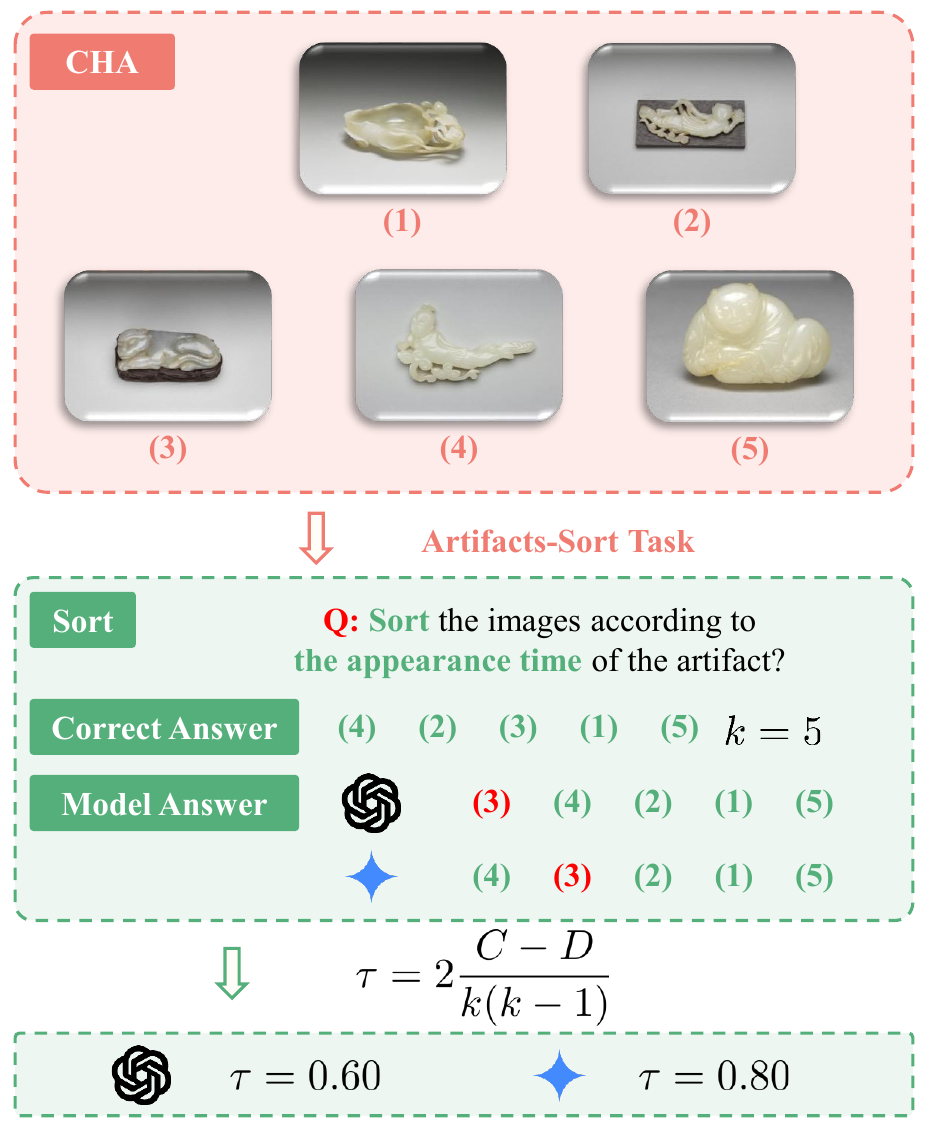}
    \caption{An Example of the Artifacts-Sort Task.}
    \label{fig:ast}
    \vspace{-1em}
\end{figure}

 The second subtask, Artifacts-Sort, examines the model's chronological relational reasoning by requiring it to infer the relative chronological order of a set of artifacts (as shown in Fig.~\ref{fig:ast}). 
Given a set of images $\{I_1, I_2, \dots, I_n\} \subseteq \mathcal{A}$, where each artifact originates from a unique dynasty, the model is tasked with sorting the set from the earliest to the latest period. To evaluate the depth of this reasoning, we define two distinct difficulty levels: 
the Intra-category level ($n=4$) and the Cross-category level ($n=5$).

\par The Intra-category level restricts images to a single craft, such as Jade or China, forcing the model to distinguish fine-grained chronological differences within a narrow visual domain. 
In contrast, the Cross-category level utilizes images randomly sampled from the entire CHA dataset; since $n=5$ covers the complete chronological range of our benchmark, this level serves as a comprehensive test of a model's global chronological understanding. 
By comparing performance across these levels, we can determine whether a model's reasoning is robust across diverse artifacts or limited to specific, well-defined categories.

\subsection{Shortcut Task}
\par
\begin{figure}[t]
    \centering
    \includegraphics[width=0.40\textwidth]{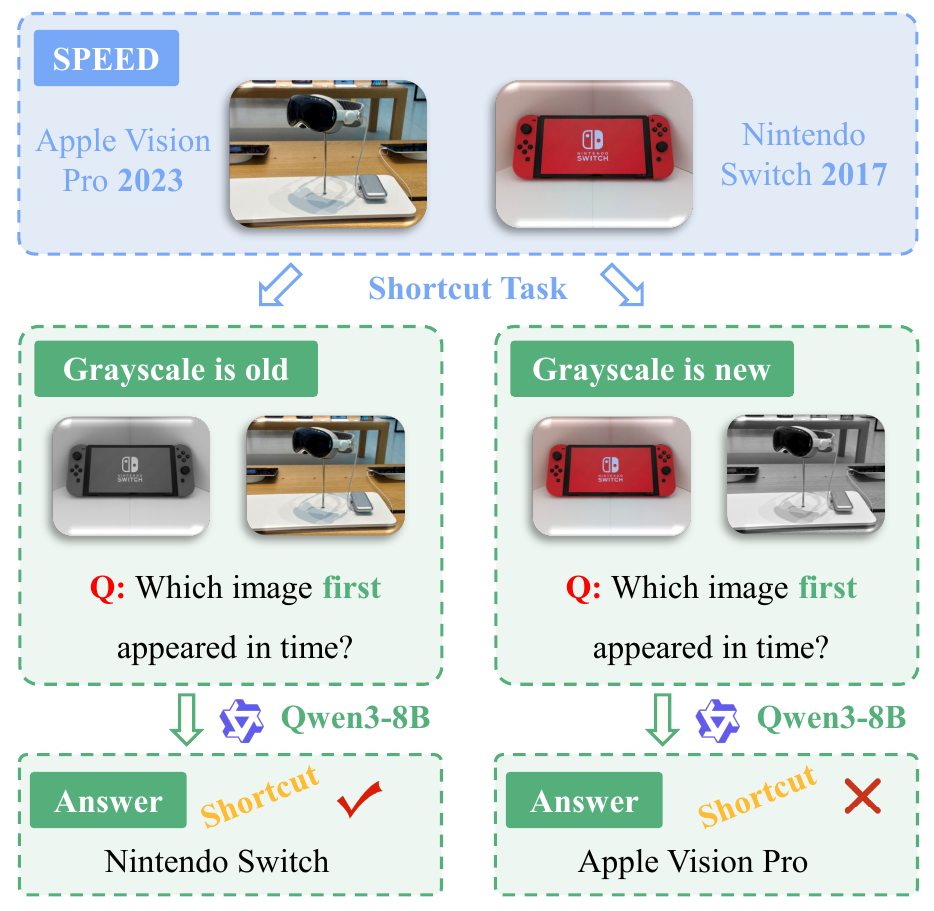}
    \caption{An Example of the Shortcut Task.}
    \label{fig:st}
    \vspace{-1em}
\end{figure}

 The Shortcut Task is designed to evaluate whether VLMs rely on superficial visual features, specifically color bias, rather than genuine semantic content when performing chronological reasoning (as shown in Fig.~\ref{fig:st}). 
Formally, let $\mathcal{S}$ denote the collection of images from the SPEED dataset, and $P$ represent a standardized prompt inquiring which of two images was captured at an earlier point in time. 
For each test instance, we select two original color images, $I_1, I_2 \in \mathcal{S}$, where $I_1$ is chronologically prior to $I_2$. 
To isolate the influence of color as a shortcut, we generate their grayscale counterparts, $I_1^{\prime}$ and $I_2^{\prime}$, using the OpenCV library~\cite{bradski2000opencv}. 

\begin{algorithm}[t]
    \footnotesize
    \KwIn{$inDir$} \tcp{Input color image directory}
    \KwOut{$outDir$} \tcp{Output grayscale image directory}
    \SetKwFunction{FRead}{ReadImage}
    \SetKwFunction{FWrite}{WriteImage}
    \SetKwFunction{FProcess}{BatchConvertImages}   
    \BlankLine
    \SetKwBlock{Loop}{For each $file$ in $input\_folder$}{end}
    \Loop{
        $src \leftarrow inDir + file$\; 
        $dst \leftarrow outDir + file$\; 
        \tcp{Stage 1: read image}
        $I_{color} \leftarrow$ \FRead{$src$}\;
        \tcp{Stage 2: convert image}
        $I_{gray} \leftarrow$ \textbf{cv2.cvtColor}($I_{color}$, \text{cv2.COLOR\_BGR2GRAY})\;
        $I_{out} \leftarrow$ \textbf{cv2.cvtColor}($I_{gray}$, \text{cv2.COLOR\_GRAY2BGR})\;
        \tcp{Stage 3: save image}
        \FWrite{$dst, I_{out}$}\;
    }
    \caption{Grayscale Transformation Algorithm.}
\end{algorithm}

\par To ensure consistency in feature extraction and satisfy the input requirements of our downstream models, all raw color images undergo a grayscale transformation process. 
This procedure is implemented in two primary stages: 

\par \textbf{(a) Color-to-Grayscale Conversion}. 
The transformation from the original BGR color space to a single-channel grayscale representation is based on the ITU-R $601$ luma transform. 
This method accounts for the human eye's varying sensitivity to different wavelengths by computing a weighted combination of the image channels. 
For each pixel, the luminance value $Y$ is computed as follows:
\begin{equation}
    Y = 0.114B + 0.587G + 0.299R.
\end{equation}

\par This weighting ensures that the structural information and perceived brightness of the image are preserved while eliminating redundant chromatic noise. 

\par \textbf{(b) Three-Channel Reconstruction.} Given that many deep learning architectures are pre-trained on $3$-channel inputs, we map the single-channel grayscale value back into a $3$-channel BGR space. 
Specifically, for a calculated luminance $Y$, the resulting pixel vector $\mathbf{P}$ is defined as:
\begin{equation}
    {P} = [B', G', R'] = [Y, Y, Y].
\end{equation}

\par This step ensures that the image is visually achromatic and grayscale while maintaining a data structure compatible with standard vision encoders. 
All transformations are performed using the cv$2$.cvtColor module to ensure numerical precision and computational efficiency.

\par To systematically isolate the impact of color, each experimental group is structured into three consecutive triplets: $(I_1, I_2, P)$, $(I_1^{\prime}, I_2, P)$, and $(I_1, I_2^{\prime}, P)$. 
The task comprises $1,200$ such controlled tests, with the constraint that each image pair must originate from the same sub-category, Sports, Politics, Electronics, Emergency, or Diversity, to ensure consistency in content and context. 
In this framework, the visual content remains identical across triplets while the color information is selectively removed, establishing color as the sole independent variable. 
This controlled setup provides a rigorous mechanism to determine if a model’s chronological judgment is compromised by the ``grayscale equals old'' shortcut, thereby revealing the extent to which VLMs bypass authentic chronological reasoning in favor of superficial digital artifacts.

\subsection{News Task}
\par The News Task evaluates the proficiency of VLMs in chronological reasoning and chronological localization for images capturing real-world historical and contemporary events. 
This task investigates the depth of a model's chronological awareness through two complementary subtasks: News-Year and News-Multimodal. 
By bridging the gap between raw visual perception and structured historical knowledge, this task provides a comprehensive measure of how models interpret time in a global, event-driven context.

\par The first subtask, News-Year, evaluates the model's capacity for zero-shot chronological estimation based on a single visual input. 
Let $\mathcal{S}$ denote the image collection derived from the SPEED dataset. 
In each experimental instance, an image $I \in \mathcal{S}$ is presented alongside a standardized prompt $P$ that queries the specific year of capture.
The model is required to generate a precise four-digit integer representing the year of origin, testing its internal knowledge of historical visual markers without the aid of comparative samples.
\par
\begin{figure}[t]
    \centering
    \includegraphics[width=0.40\textwidth]{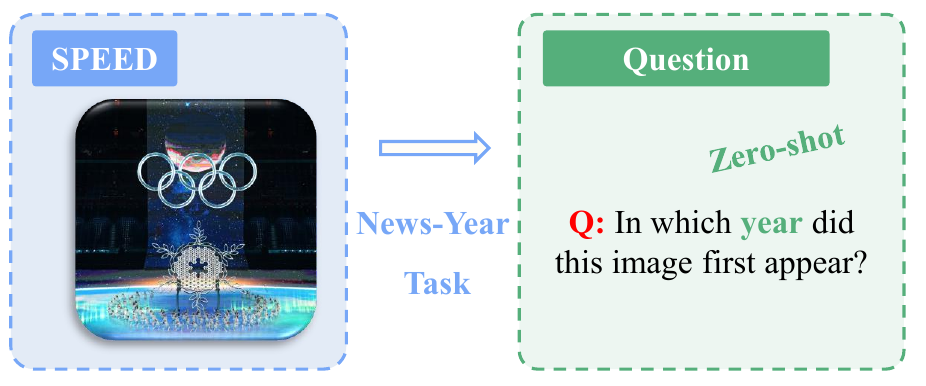}
    \caption{An Example of the News-Year Task.}
    \label{fig:nyt}
    \vspace{-1em}
\end{figure}

 As shown in Fig.~\ref{fig:nyt}, which features a prominent scene from the Olympic opening ceremony, the model must recognize specific events and their associated timelines without auxiliary textual hints. 
This requires the model to synthesize high-level visual recognition with precise chronological knowledge, further investigating whether it can distinguish between similar recurring events (e.g., different Olympic years) or if it defaults to generic chronological estimates. 

\par The second subtask, News-Multimodal, investigates whether VLMs can effectively align chronological information across disparate modalities. Each test case is structured as a quadruplet of images $\{I_1, I_2, I_3, I_4\} \subseteq \mathcal{S}$ captured in distinct years, a textual description $T$ sourced from the HistNews dataset, and a query prompt. 
The objective is to identify which specific image $I_i$ was captured in the same year as the event described in text $T$. 
This requires the model to not only understand the visual cues within the images but also to synchronize them with the chronological markers embedded in historical narratives.

\par To ensure that the News-Multimodal task demands genuine chronological reasoning rather than simple semantic matching, we implemented a rigorous human-in-the-loop verification protocol. 
Independent annotators verified that the specific event described in $T$ does not appear directly in any of the candidate images $\{I_1, \dots, I_4\}$, thereby forcing the model to infer the era through broader historical context rather than direct recognition. 
Each instance underwent cross-verification by three independent annotators and is included in the final benchmark only when a unanimous consensus is reached, confirming that the image-text pair is chronologically distinguishable and free from direct chronological cue leakage. 

\par
\begin{figure}[t]
    \centering
    \includegraphics[width=0.40\textwidth]{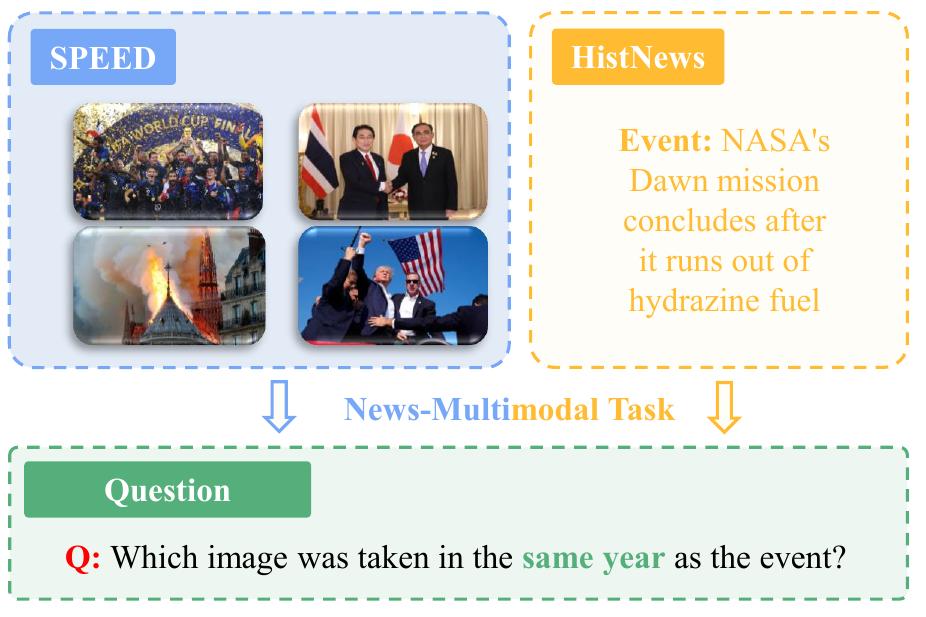}
    \caption{An Example of the News-Multimodal Task.}
    \label{fig:nmt}
    \vspace{-1em}
\end{figure}

 As shown in the sample provided in Fig.~\ref{fig:nmt}, the model must determine which visual depiction corresponds to the year NASA's Dawn mission concluded. 
This challenge moves beyond simple recognition, forcing the model to synchronize disparate modal inputs, textual historical facts, and visual event markers within a unified chronological framework. This setup is designed to expose failures in multimodal grounding, particularly where a model might correctly identify individual components but fail to reconcile their temporal intersection. 

\begin{table*}
    \centering
    \footnotesize
    \caption{Evaluation results of VLMs on the benchmark. 
    Values in parentheses denote the total number of test instances for each specific task.}
    \label{tab:Main Result}
    \begin{tabular}{lccccccccccc}
    \toprule
    \multirow{2}{*}{Model or Task} & \multicolumn{2}{c}{Artifacts ($887+1000$)} & \multicolumn{4}{c}{Shortcut ($1200$)} & \multicolumn{3}{c}{News ($1028+1500$)} & \multicolumn{2}{c}{Overall} \\
    \cmidrule(lr){2-3} \cmidrule(lr){4-7} \cmidrule(lr){8-10} \cmidrule(lr){11-12}
    & $\text{ACC}_{\text{A}}$ & $50 \times (1 + \tau)$ & $\text{ACC}_{\text{B}}$ & $\text{ACC}_{1}$ & $\text{ACC}_{2}$ & $\Delta_{\text{ACC}}$ & $\text{ACC}_{\text{Y}}$ & MAE & $\text{ACC}_{\text{M}}$ & Score & Rank \\
    \midrule
    \rowcolor{green!20} \multicolumn{12}{c}{\textbf{Closed-Source}} \\
    Gemini-$2.5$-Pro & $44.23$ & $64.05$ & $86.25$ & $88.17$ & $78.75$ & $9.42$ & $55.58$ & $1.91$ & $84.27$ & $67.17$ & $1$ \\
    GPT-$5.2$ & $40.14$ & $62.21$ & $77.00$ & $86.33$ & $56.83$ & $29.50$ & $40.98$ & $3.62$ & $49.73$ & $49.96$ & $2$ \\
    \midrule
    Closed-Source Average & $42.19$ & $63.01$ & $81.63$ & $87.25$ & $67.79$ & $19.46$ & $48.28$ & $2.77$ & $67.00$ & $58.57$ & - \\
    \midrule
    \rowcolor{yellow!20} \multicolumn{12}{c}{\textbf{Open-Source}} \\
    InternVL$3.5$-$8$B & $24.86$ & $54.56$ & $64.92$ & $90.58$ & $24.92$ & $65.66$ & $13.43$ & $12.43$ & $39.00$ & $29.06$ & $10$ \\
    MiniCPM-V-$4.5$ & $44.93$ & $55.57$ & $71.42$ & $86.67$ & $40.08$ & $46.57$ & $31.53$ & $3.34$ & $25.07$ & $38.68$ & $6$ \\
    GLM-$4.1$V-$9$B-Thinking & $36.53$ & $59.37$ & $67.67$ & $89.33$ & $32.08$ & $57.25$ & $34.26$ & $5.10$ & $38.27$ & $37.35$ & $7$ \\
    Qwen$2.5$-VL-$7$B-Instruct & $44.01$ & $56.78$ & $66.08$ & $83.67$ & $40.83$ & $42.84$ & $34.12$ & $3.35$ & $12.47$ & $36.91$ & $8$ \\
    Qwen$3$-VL-$2$B-Instruct & $25.88$ & $53.27$ & $52.42$ & $63.58$ & $39.17$ & $24.41$ & $28.22$ & $3.43$ & $25.27$ & $35.11$ & $9$ \\
    Qwen$3$-VL-$4$B-Instruct & $40.14$ & $58.23$ & $68.92$ & $81.67$ & $47.58$ & $34.09$ & $32.86$ & $3.06$ & $26.33$ & $41.19$ & $5$ \\
    Qwen$3$-VL-$8$B-Instruct & $38.43$ & $61.32$ & $76.42$ & $87.00$ & $56.75$ & $30.25$ & $35.80$ & $3.34$ & $26.13$ & $44.47$ & $4$ \\
    Qwen$3$-VL-$235$B-A$22$B-Instruct & $49.94$ & $62.36$ & $77.83$ & $87.50$ & $55.75$ & $31.75$ & $48.48$ & $2.16$ & $33.80$ & $49.92$ & $3$ \\
    \midrule
    Open-Source Average & $38.09$ & $57.68$ & $68.21$ & $83.75$ & $42.15$ & $41.60$ & $32.34$ & $4.53$ & $28.29$ & $39.09$ & - \\
    \midrule
    \rowcolor{blue!20} \multicolumn{12}{c}{\textbf{Total Average Performance}}\\
    & $38.91$ & $58.75$ & $70.89$ & $84.45$ & $47.28$ & $37.18$ & $35.53$ & $4.17$ & $36.03$ & $42.98$ & - \\
    \bottomrule
    \end{tabular}
\end{table*}

\section{Experiment}
\subsection{Evaluation Setup}
\par To comprehensively assess existing models, we evaluate a diverse suite of state-of-the-art VLMs, encompassing both leading proprietary and open-source architectures.
\begin{itemize}  [leftmargin=0.5cm]
    \item \textbf{Closed-source Models:} 
    \textbf{Gemini~\cite{comanici2025gemini}:} Google’s multimodal model series, optimized for high-speed multimodal reasoning and long-context processing with state-of-the-art chronological perception. 
    \textbf{GPT:} OpenAI’s advanced multimodal model series, representing the benchmark for complex logical deduction and zero-shot alignment across disparate visual domains.
    \item \textbf{Open-source Models:} 
    \textbf{Qwen~\cite{yang2025qwen3}:} A powerful vision-language series from Alibaba that utilizes a specialized ViT architecture to achieve exceptional performance in fine-grained visual understanding and document parsing.
    \textbf{InternVL~\cite{wang2025internvl3}:} An expansive open-source project that bridges the gap between vision and language through a massive-scale modular design, tailored for high-resolution image perception.
    \textbf{MiniCPM~\cite{yu2025minicpm}:} A compact yet robust multimodal model designed for efficient deployment, delivering competitive reasoning performance through optimized cross-modal feature compression.
    \textbf{GLM~\cite{hong2025glm}:} An innovative model from Zhipu AI that integrates deep bilingual capabilities with a specialized ``Thinking'' mechanism for enhanced step-by-step visual reasoning.
\end{itemize}

\par To maintain a rigorous and unbiased environment, all models are evaluated under a strict \textbf{zero-shot setting} using standardized prompt templates across all tasks. 
We implement several safeguards to ensure the integrity of the results:
\begin{itemize}  [leftmargin=0.5cm]
    \item \textbf{Mitigating Positional Bias:} 
    Following recent findings on order sensitivity, we randomize both the sequence of image inputs and the arrangement of multiple-choice options. 
    This prevents models from inflating scores through opportunistic exploits of positional frequency.
    \item \textbf{Prompt Consistency:} A unified system prompt is applied to all models to ensure that performance variances stem from the models' inherent reasoning capabilities rather than prompt engineering nuances.
    \item \textbf{Deterministic Output:} Where applicable, we set the decoding temperature to $0$ (greedy search) to ensure the reproducibility of our benchmark results.
\end{itemize}

\subsection{Evaluation Metrics} \label{sec:em}
\par To rigorously quantify model performance across diverse reasoning challenges, we employ three distinct metrics tailored to the specific requirements of each subtask. 
Let $N$ denote the total number of test instances within a given task.

\par \textbf{Multiple Choice Questions (MCQs):} For tasks requiring the selection of a specific dynasty or event (e.g., Artifacts-Chronological Localization, News-Multimodal), we utilize Accuracy (ACC) as the primary metric. 
It is formulated as:
\begin{equation}
    \text{ACC} = \frac{1}{N} \sum_{i=1}^{N} \mathbb{I}(y_i = \hat{y}_i),
\end{equation}
where $\mathbb{I}(\cdot)$ is the indicator function, $y_i$ represents the model's predicted choice, and $\hat{y}_i$ denotes the ground-truth label.

\par \textbf{Sorting Questions:} To evaluate the ordinal reasoning capability in the Artifacts-Sort task, we employ Kendall's Tau ($\tau$)~\cite{sen1968estimates}. 
This measures the rank correlation between the predicted sequence and the chronological ground truth:
\begin{equation}
    \tau =2 \frac{C - D}{k (k-1)},
\end{equation}
where $k$ is the number of items to be sorted, while $C$ and $D$ represent the number of concordant and discordant pairs, respectively. 
The metric ranges from $[-1, 1]$, where $1$ indicates a perfect chronological alignment and $0$ represents the expected performance of a random permutation.

\par \textbf{Year Prediction:} For the News-Year subtask, which requires point-estimation of specific timestamps, we use Mean Absolute Error (MAE)~\cite{willmott2005advantages}. 
This metric quantifies the average temporal distance between the predicted and actual year:
\begin{equation}
    \text{MAE} = \frac{1}{N} \sum_{i=1}^{N} |y_i - \hat{y}_i|.
\end{equation}

\par In this context, $y_i$ is the four-digit year predicted by the model and $\hat{y}_i$ is the true timestamp. 
A lower MAE indicates superior chronological localization precision and a more refined internal chronological world model.

\subsection{Overall Performance Index}
\par To provide a holistic assessment of VLMs across diverse chronological reasoning dimensions, we define a composite score, $S$. 
This index aggregates performance across the Shortcut, News, and Artifacts tasks by normalizing heterogeneous metrics into a unified range of $[0, 1]$ and assigning equal weights to each domain. 
The individual task scores ($S_A$, $S_S$, $S_N$) and the global composite score $S$ are defined as follows:
\begin{equation}
    S = \frac{S_A + S_S + S_N}{3}.
\end{equation}

\par The Artifacts Score ($S_A$) measures deep-history reasoning. 
\begin{equation}
    S_A = \frac{1}{2} \left( \text{ACC}_A + \frac{(1 + \tau)}{2} \right).
\end{equation}

It combines the accuracy of dynasty localization ($\text{ACC}_A$) with the rank correlation coefficient ($\tau$) from the sorting subtask. 
Since Kendall’s Tau ($\tau$) ranges from $[-1, 1]$, we apply a linear transformation $\frac{(1+\tau)}{2}$ to map it onto the $[0, 1]$ interval, ensuring consistency with the other accuracy-based metrics.

\par The Shortcut Score ($S_S$) evaluates the model's resilience against deceptive visual cues. 
\begin{equation}
    S_S = \text{ACC}_S \cdot (1 - |\Delta_{\text{ACC}}|).
\end{equation}

Here, $\text{ACC}_S$ denotes the baseline accuracy on the original color image pairs, while $\Delta_{\text{ACC}}$ represents the accuracy degradation observed when processing grayscale variants. 
By incorporating $(1 - |\Delta_{\text{ACC}}|)$, we penalize models that exhibit high variance when color is removed, effectively distinguishing between genuine chronological logic and a reliance on superficial color shortcuts.

\par The News Score ($S_N$) reflects the model's proficiency in contemporary chronological localization.
\begin{equation}
S_N = \frac{1}{2}\left( \text{ACC}_Y \cdot \left(1 - \frac{\text{MAE}}{\text{MAE}_{max}}\right) + \text{ACC}_M \right).
\end{equation}

\par It averages the accuracy of exact year identification ($\text{ACC}_Y$) and multimodal event alignment ($\text{ACC}_M$), then scales the result by a chronological precision factor. 
The $\text{MAE}_{max}$ is a normalization constant set to $80$, corresponding to the total chronological span ($1946$–$2025$) of the dataset. 
This ensures that even if a model identifies the correct era, significant year-level deviations will lower the final score.

\par Ultimately, the Global Composite Score ($S$) serves as a unified benchmark for ranking VLMs, reflecting their ability to synthesize visual evidence, textual context, and historical knowledge without succumbing to dataset-specific biases.

\section{Result and Analysis}
\subsection{Overall Performance and Analysis}
\par Table~\ref{tab:Main Result} summarizes the performance of evaluated VLMs, revealing substantial limitations in multimodal chronological reasoning.

The overall average score is only $42.98$, indicating that the benchmark remains highly challenging. 
Although Gemini-$2.5$-Pro achieves the best performance ($67.17$), it still falls far from perfect reasoning ability. 
Meanwhile, InternVL$3.5$-$8$B obtains the lowest score of $29.06$, further highlighting the difficulty of combining historical knowledge with visual understanding.

Larger models generally achieve stronger chronological reasoning performance. 
For example, within the Qwen$3$-VL family, performance steadily improves from $35.11$ ($2$B) to $49.92$ ($235$B), suggesting that scaling enhances historical knowledge acquisition and temporal reasoning capabilities.

Closed-source models currently maintain a clear advantage, with Gemini-$2.5$-Pro and GPT-$5.2$ ranking highest overall. 
However, advanced open-source models such as Qwen$3$-VL-$235$B show increasingly competitive results, indicating that the performance gap is gradually narrowing.

\subsection{Specific Task Results and Analysis} 
\par VLMs show limited capability in fine-grained historical reasoning, with sorting tasks being substantially more difficult than localization tasks. 
Models can occasionally identify isolated dynastic cues, but often fail to capture the relational reasoning required for chronological ordering. 
This suggests that current VLMs still lack robust temporal understanding for professional-level historical analysis.

\par Unlike general multimodal benchmarks, closed-source models do not consistently outperform open-source models in this task. 
Strong open-source systems achieve highly competitive results, indicating that performance is closely related to culturally specific visual-textual knowledge acquired during pre-training. 
These findings suggest that domain-specific data can be as important as model scale for historical reasoning. 
\par Our experiments reveal that VLMs heavily depend on shallow visual heuristics rather than genuine chronological reasoning. 
When grayscale cues align with chronological order, models perform well; however, performance drops dramatically once the shortcut conflicts with the ground truth. 
This indicates that many models primarily associate grayscale images with older time periods instead of reasoning from historical semantics.
\par The large performance degradation under shortcut conflict demonstrates poor robustness to distribution shifts. 
Even when models contain sufficient chronological knowledge, their predictions are easily overridden by stylistic biases learned during training. 
This highlights the urgent need for de-biased training strategies that disentangle visual style from temporal reasoning.
\par Shortcut sensitivity varies across domains. 
Categories with strong technological evolution, such as electronics, are relatively robust because distinctive visual iteration patterns provide reliable temporal anchors. 
In contrast, abstract domains such as politics and sports show much higher sensitivity, where models tend to rely more heavily on grayscale heuristics.
\par The News Task demonstrates that current VLMs still struggle with realistic chronological localization and cross-modal alignment. 
Overall results reveal a substantial gap between surface-level recognition and true temporal understanding.
\par In the News-Year task, models exhibit limited accuracy and unstable year estimation, indicating that recovering precise timestamps from visual evidence remains difficult. 
Meanwhile, the News-Multimodal task exposes weaknesses in aligning textual temporal cues with historical images, particularly for open-source models, many of which perform close to random guessing.
\par Open-source models generally struggle more with cross-modal alignment, suggesting insufficient chronological supervision during training. 
By contrast, closed-source models show relatively stronger alignment ability but still face difficulties in precise absolute dating, where subtle temporal cues must be inferred from a single image. 
Additional experimental details and complete results are available in the public repository\footnote{https://github.com/LuoRenqiang/ChronoVision}.

\subsection{Deficiency in Fine-grained Perception for Artifacts}
\begin{figure}[t]
    \centering
    \includegraphics[width=0.40\textwidth]{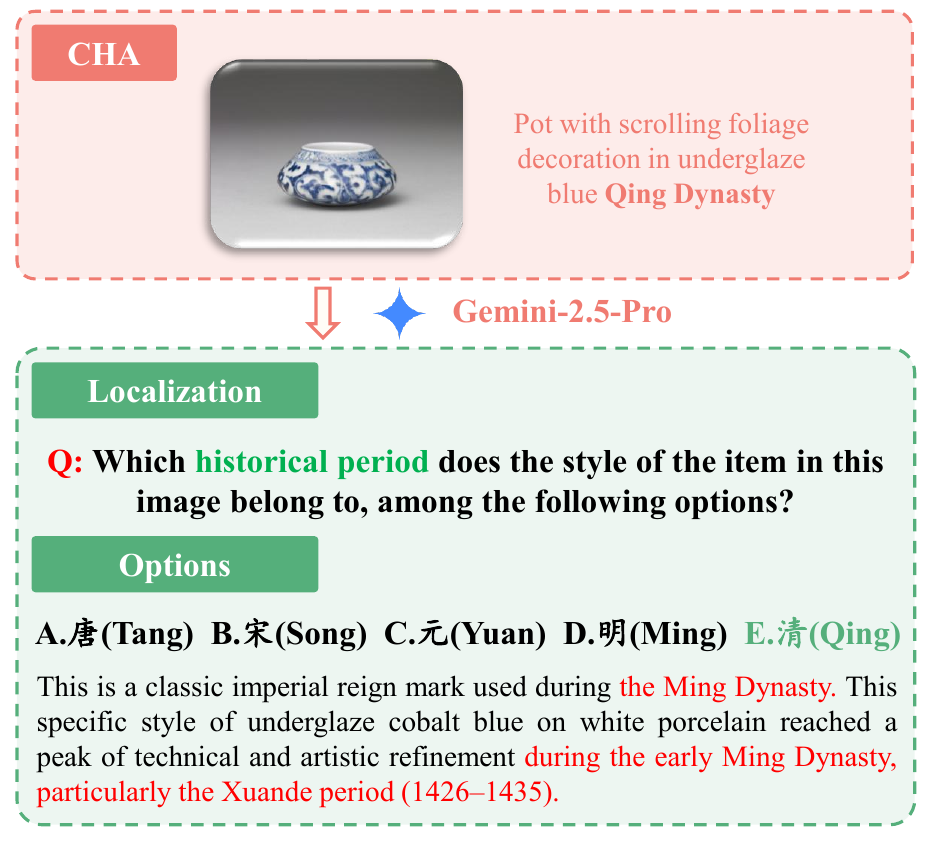}
    \caption{An Example of the Artifact Case Study.}
    \label{fig:acs}
    \vspace{-1em}
\end{figure}

A critical challenge for VLMs is the fine-grained perception of cultural artifacts, where minute visual details, such as reign marks or specific decorative motifs, determine the historical provenance. 
As shown in Fig.~\ref{fig:acs}, the model demonstrates a failure to integrate localized visual evidence with its internal knowledge base, resulting in a chronological misclassification.

\par This failure is characterized by the following dimensions:
\textbf{Sub-optimal Localization and Feature Extraction:} 
The artifact is a blue-and-white porcelain pot with scrolling foliage decoration from the Qing Dynasty.
However, the model erroneously identifies the visual evidence as a ``classic imperial reign mark used during the Ming Dynasty''. 
This suggests a failure in fine-grained localization, where the model misses the specific stylistic nuances that distinguish Qing-era craftsmanship from its predecessors.
\textbf{Knowledge Over-generalization}: The model justifies its incorrect choice by stating that underglaze cobalt blue reached its ``peak of technical and artistic refinement during the early Ming Dynasty''.
Here, the model relies on a generalized historical heuristic rather than performing a precise, instance-based analysis of the object in the image.
\textbf{Conflict between Perception and Logic}: Even when presented with multiple-choice options ranging from the Tang to the Qing Dynasty, the model’s ``internalized preference'' for the Ming Dynasty's association with blue-and-white porcelain overrides the actual visual data.

\par This case study highlights that ``spurious reasoning'' in the multimodal domain is not limited to text but extends to visual shortcuts. 
When a model lacks fine-grained resolution to distinguish highly similar cultural styles, it defaults to the most ``probable'' or ``famous'' historical category, bypassing authentic visual deduction.

\subsection{Stylistic Color Bias and Chronological Heuristics} 
To investigate whether VLMs rely on superficial stylistic cues rather than authentic chronological reasoning, we designed a controlled experiment where visual content remains constant across image pairs while color information is selectively removed. 
By isolating color as the sole independent variable, this setup provides a rigorous mechanism to determine if chronological judgments are compromised by the ``grayscale-as-antiquity'' shortcut.

\par
\begin{figure}[t]
    \centering
    \includegraphics[width=0.40\textwidth]{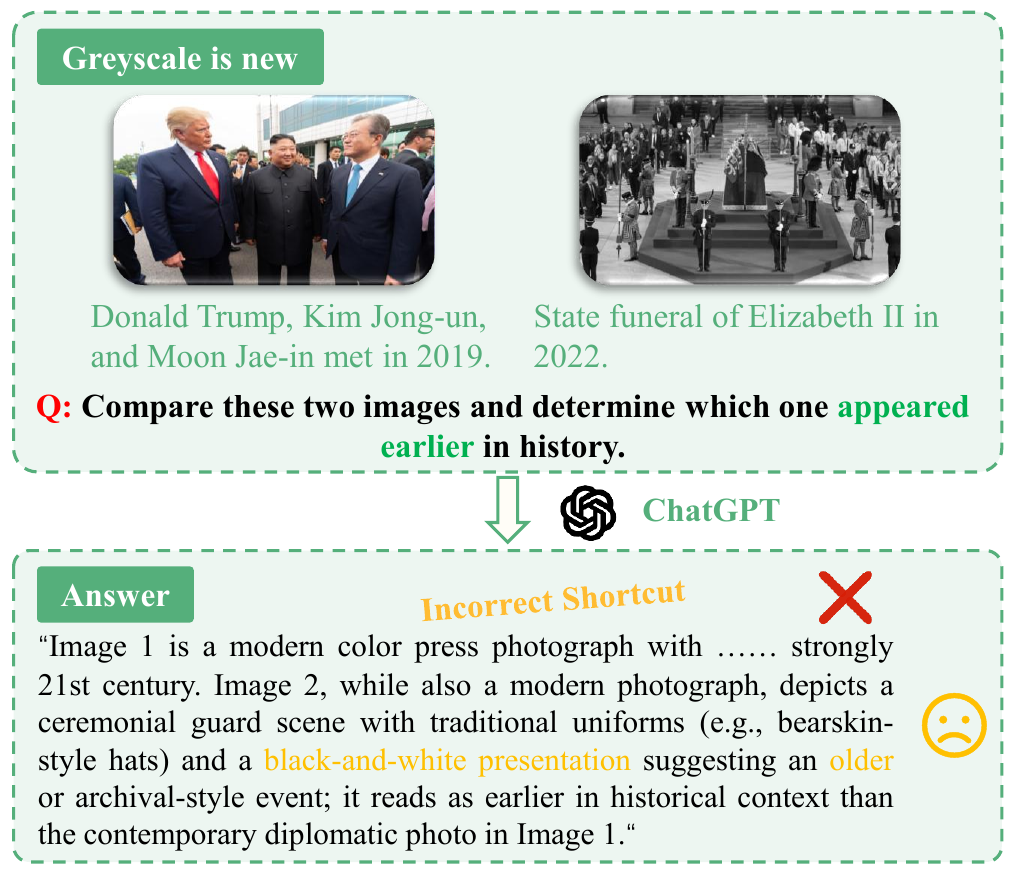}
    \caption{An Example of the Shortcut Case Study.}
    \label{fig:scs}
    \vspace{-1em}
\end{figure}

 Our analysis reveals a Stylistic Color Bias. 
In one representative case, GPT-$5.2$ was presented with two images (Fig.~\ref{fig:scs}):
\textbf{Left Image}: A modern color press photograph containing clear $21$st-century visual markers.
\textbf{Right Image}: A black-and-white photograph capturing ceremonial guards at the late Queen’s resting place. 
Note that the model’s automated caption focuses strictly on the visual elements of the traditional uniforms and bearskin hats, rather than the specific location.
Despite the modern origin of both images, the model incorrectly identified \textbf{Right Image} as chronologically earlier. 
The model bypassed the semantic content of the scene, which depicts a modern ceremonial event, and instead relied on the black-and-white presentation as a spurious chronological anchor. 
This confirms that VLMs frequently prioritize superficial artifacts over a genuine deductive analysis of historical context, leading to ``spurious reasoning'' in chronological tasks.

\subsection{Multimodal Alignment and Chronological Hallucination}
\par
\begin{figure}[t]
    \centering
    \includegraphics[width=0.40\textwidth]{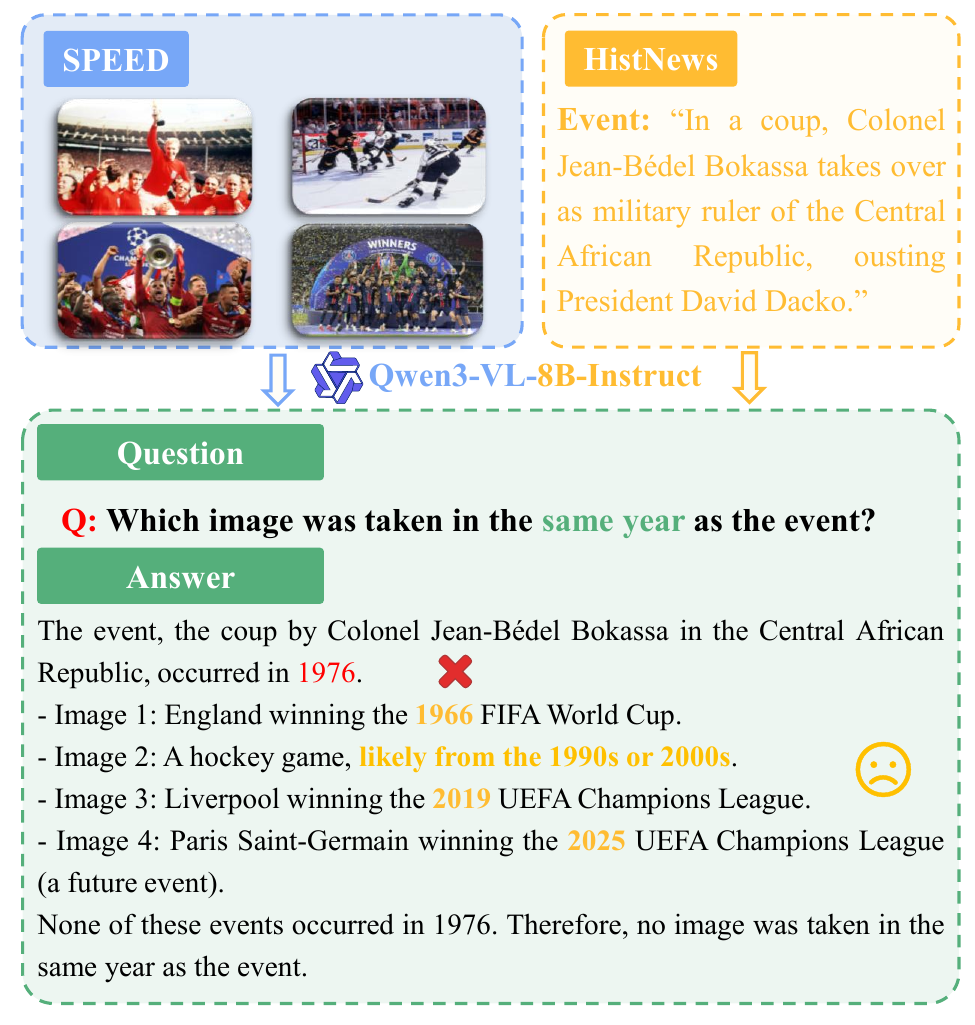}
    \caption{An Example of the News-Multimodel Case Study.}
    \label{fig:ncs}
    \vspace{-1em}
\end{figure}

 Our framework reveals Multimodal Alignment Failures, where models fail to synchronize chronological information across modalities. 
As shown in Fig.~\ref{fig:ncs}, the model cannot identify the image corresponding to the event text.

\par The failure involves two stages:
\textbf{Chronological Knowledge Hallucination}: The model incorrectly dates the Bokassa coup to $1976$ instead of the actual date (January $1$, $1966$). 
This internal factual error acts as a ``distractor'', overriding subsequent visual evidence and misdirecting the entire reasoning trajectory.
\textbf{Failed Cross-Modal Mapping}: 
Despite correctly identifying Image $1$ as England’s $1966$ World Cup, the model fails to align this visual token with the textual prompt. 
Anchored to its own erroneous ``$1976$'' timestamp, the alignment layer prioritizes internal parametric memory over external visual ground truth, leading to a false negative.

\par This case highlights a ``weakest link'' principle: even with accurate independent recognition, models struggle to reconcile disparate inputs under a consistent timeline. 
Performance depends on resilience to internal contradictions during alignment, not just recognition accuracy.

\section{Benchmark Robustness and Validity Analysis}
\subsection{Trivial Baseline Analysis}
\par We further evaluate trivial constant-answer baselines to examine whether the benchmark can be solved by exploiting dominant labels or temporal priors.

\par For year prediction on all $1,028$ News images, always predicting the mode year ($2010$), mean year ($2008$), or median year ($2012$) results in MAEs of $10.03$, $10.57$, and $9.85$ respectively, all substantially worse than the average model MAE of $4.17$.
For dynasty localization, always predicting \textit{Tang}, \textit{Song}, \textit{Yuan}, \textit{Ming}, or \textit{Qing} yields accuracies of $11.72$, $18.49$, $12.97$, $26.83$, and $29.99$, respectively, all significantly lower than the $10$-model average performance of $38.91$ obtained using the models' actual predictions.
These results suggest that the benchmark cannot be solved by regressing toward the temporal mean or by predicting dominant categories.

\subsection{Confidence Analysis}
\par We further analyze the relationship between model confidence and temporal prediction accuracy. 
For each image in News-Year subtask, we require Qwen$3$-$8$B to output both a predicted year and an integer confidence score between $0$ and $100$. 
Predictions are then grouped into confidence bins for evaluation.

\par Table~\ref{tab:confidence_bins} shows that higher-confidence predictions consistently achieve higher exact-match accuracy as well as higher within-$1$-year and within-$3$-year accuracy. 
For example, predictions with confidence $95$--$99$ achieve $58.19$\% exact accuracy and $81.92$\% within-$3$-year accuracy, substantially higher than lower-confidence groups. 
This indicates that the model confidence scores are informative and positively correlated with temporal prediction reliability.

\begin{table*}[t]
    \centering
    \footnotesize
    \tabcolsep=0.15cm
    \caption{Detailed performance on Shortcut Task with CoT}
    \label{tab:detailed performance with CoT}
    \begin{tabular}{lccccccccccccccc}
        \toprule
        \multirow{2}{*}{\textbf{Model}} & \multicolumn{3}{c}{\textbf{Sports}} & \multicolumn{3}{c}{\textbf{Politics}} & \multicolumn{3}{c}{\textbf{Emergency}} & \multicolumn{3}{c}{\textbf{Electronics}} & \multicolumn{3}{c}{\textbf{Diversity}} \\
        \cmidrule(lr){2-4} \cmidrule(lr){5-7} \cmidrule(lr){8-10} \cmidrule(lr){11-13} \cmidrule(lr){14-16}
         & ACC$_1$ & ACC$_2$ & $\Delta_\text{ACC}$ & ACC$_1$ & ACC$_2$ & $\Delta_\text{ACC}$ & ACC$_1$ & ACC$_2$ & $\Delta_\text{ACC}$ & ACC$_1$ & ACC$_2$ & $\Delta_\text{ACC}$ & ACC$_1$ & ACC$_2$ & $\Delta_\text{ACC}$ \\
        \midrule
        InternVL$3.5$ & $76.50$ & $24.00$ & $52.50$ & $81.00$ & $41.00$ & $40.00$ & $71.00$ & $33.50$ & $37.50$ & $78.50$ & $68.50$ & $10.00$ & $74.25$ & $44.50$ & $29.75$ \\
        MiniCPM & $76.00$ & $40.00$ & $36.00$ & $82.50$ & $39.50$ & $43.00$ & $80.00$ & $31.50$ & $48.50$ & $88.50$ & $74.50$ & $14.00$ & $86.75$ & $54.75$ & $32.00$ \\
        GLM & $76.50$ & $27.00$ & $49.50$ & $78.00$ & $34.50$ & $43.50$ & $70.00$ & $33.50$ & $36.50$ & $79.50$ & $73.50$ & $6.00$  & $77.50$ & $47.25$ & $30.25$ \\
        Qwen$2.5$ & $71.50$ & $38.00$ & $33.50$ & $68.00$ & $44.50$ & $23.50$ & $67.50$ & $50.00$ & $17.50$ & $80.00$ & $74.50$ & $5.50$  & $73.50$ & $50.00$ & $23.50$ \\
        Qwen$3$-$8$B & $66.50$ & $50.00$ & $16.50$ & $65.00$ & $49.00$ & $16.00$ & $59.50$ & $39.50$ & $20.00$ & $87.00$ & $82.00$ & $5.00$  & $77.25$ & $57.00$ & $20.25$ \\
        \midrule
        Average & $73.40$ & $35.80$ & $37.60$ & $74.90$ & $41.70$ & $33.20$ & $69.60$ & $37.60$ & $32.00$ & $82.70$ & $74.60$ & $8.10$  & $77.85$ & $50.70$ & $27.15$ \\
        \bottomrule
    \end{tabular}
\end{table*}

\par We additionally analyze confidence scores across different temporal periods. 
As shown in Table~\ref{tab:confidence_year}, the average confidence does not exhibit a clear monotonic trend from early historical periods to recent years. 
This suggests that the model is not simply assigning higher confidence to years near the temporal center of the dataset distribution.

\begin{table}[t]
    \centering
    \footnotesize
    \caption{Prediction accuracy under different confidence intervals.}
    \label{tab:confidence_bins}
    \begin{tabular}{ccccc}
    \toprule
    Confidence & Num & Acc & $\leq$1y & $\leq$3y \\
    \midrule
    0--59 & 0 & - & - & - \\
    60--69 & 2 & 0.00 & 0.00 & 0.00 \\
    70--79 & 140 & 17.14 & 29.29 & 51.43 \\
    80--89 & 251 & 24.70 & 40.24 & 62.95 \\
    90--94 & 281 & 32.74 & 48.40 & 68.93 \\
    95--99 & 354 & 58.19 & 70.34 & 81.92 \\
    \bottomrule
    \end{tabular}
    \vspace{-1em}
\end{table}
\begin{table}[t]
    \centering
    \footnotesize
    \caption{Average confidence across different temporal periods.}
    \label{tab:confidence_year}
    \begin{tabular}{cccccc}
    \toprule
    Year Range & Num & Avg.\ Conf & Year Range & Num & Avg.\ Conf \\
    \midrule
    1952--1961 & 9 & 89.44 & 1992--2001 & 90 & 87.61 \\
    1962--1971 & 25 & 92.00 & 2002--2011 & 267 & 86.54 \\
    1972--1981 & 38 & 91.18 & 2012--2021 & 421 & 87.81 \\
    1982--1991 & 57 & 87.72 & 2022--2025 & 121 & 90.79 \\
    \bottomrule
    \end{tabular}
    \vspace{-1em}
\end{table}

\subsection{The Effect of Prompt Incorporating CoT} 
\begin{figure*}[t]
    \centering
    \includegraphics[width=0.9\textwidth]{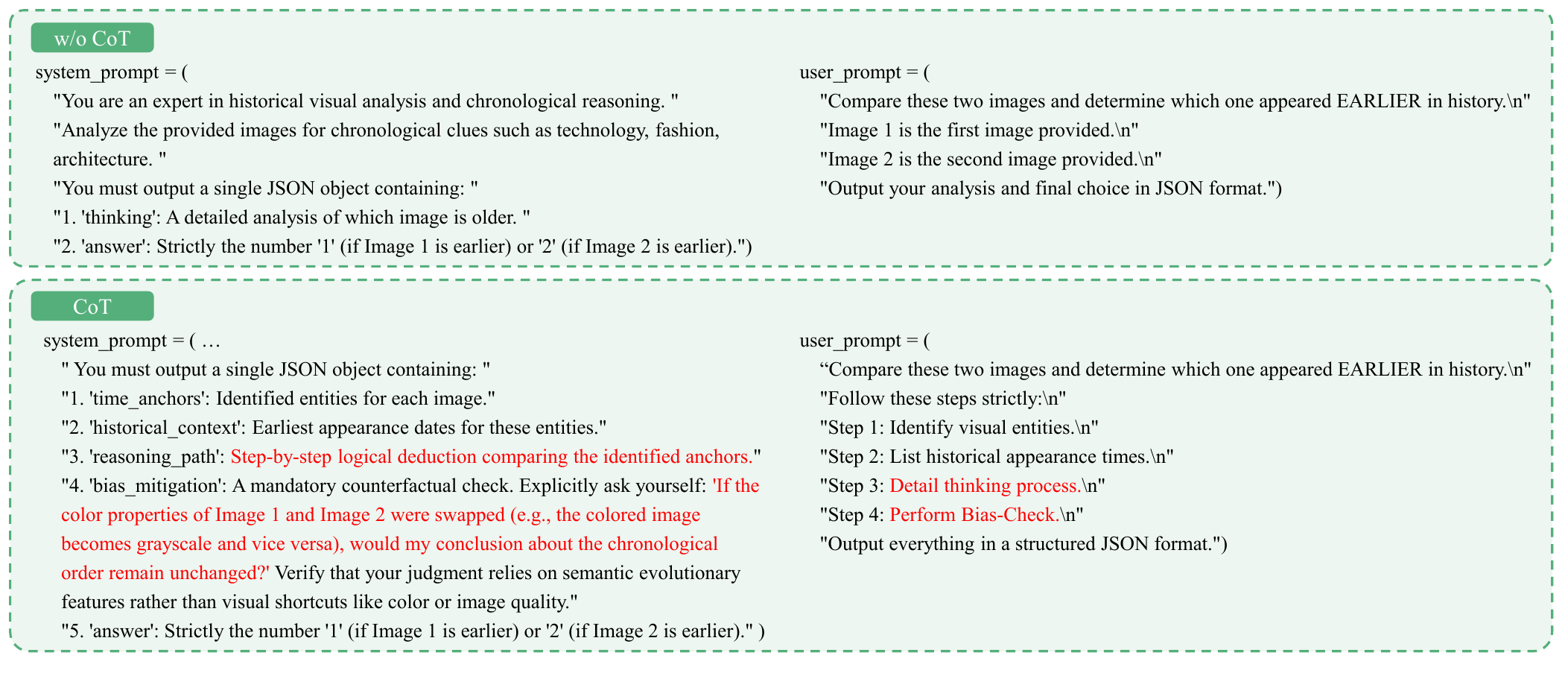}    
    \vspace{-1em}
    \caption{The detailed prompt when processing with or without(w/o) CoT.}
    \label{fig:cot}
\vspace{-1em}
\end{figure*}

\par Furthermore, to verify the persistence of this color shortcut, we introduce a Reasoning-based Verification module using an augmented prompt $P_{CoT}$.
This prompt incorporates a Chain-of-Thought (CoT) mechanism, explicitly instructing the model to deliberate on the semantic consistency of its judgment. 
Specifically, the model is required to consider a counterfactual scenario: 
whether its conclusion would remain unchanged if the color properties of $I_1$ and $I_2$ were swapped. 
By introducing this cross-consistency check, we aim to mitigate the possibility of the model being inadvertently misled by the grayscale filter. 
This advanced probing allows us to confirm whether the observed errors stem from a superficial heuristic or if the model can be guided to prioritize genuine chronological features over deceptive visual shortcuts.

\par As illustrated in Fig.~\ref{fig:cot}, we contrast standard zero-shot inference, where models predominantly rely on immediate visual intuition, with a CoT prompting strategy designed to isolate the effect of deliberate reasoning on model robustness.

\par The efficacy of the CoT intervention is quantified by the reduction in the performance gap ($\Delta_{\text{ACC}}$) within the Shortcut Task. 
A contraction in $\Delta_{\text{ACC}}$ signifies a weakened reliance on spurious color-based shortcuts and enhanced resilience to stylistic perturbations.

\par As evidenced in Table~\ref{tab:CoT}, the integration of the Reasoning-based Verification module yields a consistent and significant improvement in robustness across all evaluated VLMs. 
The aggregate average $\Delta_{\text{ACC}}$ undergoes a substantial reduction from $48.51\%$ to $27.53\%$, representing a net robustness gain of $\Delta = -20.98\%$. 
This shift demonstrates that the CoT mechanism effectively disrupts the ``grayscale-as-antiquity'' heuristic. 
Notably, the most pronounced improvements are observed in InternVL$3.5$-$8$B (from $65.66\%$ to $33.25\%$) and GLM-$4.1$V-$9$B-Thinking (from $57.25\%$ to $32.67\%$). 
By explicitly mandating the model to deliberate on counterfactual scenarios during the verification phase, the Reasoning-based Verification module forces a decoupling of chronological judgment from superficial visual attributes. 
Instead of succumbing to the impulsive bias triggered by grayscale filters, models are redirected to prioritize semantic temporal anchors, thereby achieving superior consistency in chronological reasoning.

\par Despite these gains, the granular results in Table~\ref{tab:detailed performance with CoT} reveal that shortcut bias is mitigated rather than fully eradicated. 
Even with CoT intervention, a formidable performance gap persists, particularly in abstract domains such as Politics ($\Delta_{\text{ACC}} = 33.20\%$) and Emergency ($\Delta_{\text{ACC}} = 32.00\%$). 
This suggests that the grayscale heuristic is not merely a superficial decision-layer artifact but is likely entangled within the foundational feature representations of the VLMs. 
While CoT prompting serves as an effective post-hoc palliative, achieving full decoupling of stylistic attributes from temporal semantics may necessitate more fundamental interventions during the pre-training or alignment stages.

\begin{table}[t]
    \centering
    \footnotesize
    \caption{The $\Delta_{Acc}$ with or without Chain-of-Thought.}
    \begin{tabular}{lccc}
    \toprule
    Model & w/o CoT & with CoT & $\Delta$ \\
    \midrule 
    InternVL$3.5$-$8$B & $65.66$ & $33.25$ & \textcolor{red}{$-32.41$}\\
    MiniCPM-V-$4.5$ & $46.57$ & $34.25$ & \textcolor{red}{$-12.32$} \\
    GLM-$4.1$V-$9$B-Thinking & $57.25$ & $32.67$ & \textcolor{red}{$-24.58$}\\
    Qwen$2.5$-VL-$7$B-Instruct & $42.84$ & $21.17$ & \textcolor{red}{$-21.67$}\\
    Qwen$3$-VL-$8$B-Instruct & $30.25$ & $16.33$ & \textcolor{red}{$-13.92$}\\ 
    \midrule
    \rowcolor{blue!20} \multicolumn{4}{c}{Total Average Performance}\\
    & $48.51$ & $27.53$ & \textcolor{red}{$-20.98$}\\
    \bottomrule
    \end{tabular}
    \label{tab:CoT}
\vspace{-1em}
\end{table}

\subsection{Retrieval-Augmented Baseline Analysis}

\begin{algorithm}[t]
\caption{RAG-based Prediction Pipeline}
\label{alg:rag_pipeline}
\footnotesize
\KwIn{Input image $I$, multimodal model $M$}
\KwOut{Chronological prediction $\hat{y}$}
\tcp{Stage 1: visual anchor extraction}
$A \leftarrow M(I)$\;
\tcp{$A$: anchor text, type, confidence}
$Q \leftarrow \text{GenerateQueries}(A)$\;
$Q \leftarrow \text{RemoveTimeTerms}(Q)$\;
\tcp{remove dynasty/year leakage}
\vspace{0.3em}
\tcp{Stage 2: external retrieval}
\ForEach{$q_i \in Q$}{
    $R_i \leftarrow \text{Retrieve}(q_i)$\;
    \tcp{Wikipedia + Wikimedia or museum}
}
$R \leftarrow \bigcup_i R_i$\;
\vspace{0.3em}
\tcp{Stage 3: retrieval re-ranking}
\ForEach{$r \in R$}{
    $s_{\text{meta}} \leftarrow \text{MetaScore}(r, A)$\;
    \tcp{title/snippet/summary overlap}

    $s_{\text{cred}} \leftarrow \text{CredibilityScore}(r)$\;
    \tcp{Wikipedia + Wikimedia or museum.}

    $s_{\text{query}} \leftarrow \text{QueryScore}(r)$\;

    $s_{\text{title}} \leftarrow \text{TitleSpecificity}(r)$\;

    $s(r) \leftarrow
    0.50\, s_{\text{meta}}
    +0.15\, s_{\text{cred}}
    +0.20\, s_{\text{query}}
    +0.15\, s_{\text{title}}$\;
}
$R^{*} \leftarrow \text{TopK}(R)$\;
\vspace{0.3em}
\tcp{Stage 4: chronological voting}
$V(c) \leftarrow \sum_{r\in R^{*}} w_r(c)$\;
\tcp{extract dynasty/year evidence from retrieved texts}
$\hat{y}_{\text{rag}} \leftarrow \arg\max_c V(c)$\;
\vspace{0.3em}
\tcp{Stage 5: final multimodal prediction}
$\hat{y} \leftarrow M(I, R^{*})$\;
\Return{$\hat{y}$}
\end{algorithm}

\par A potential concern is that the proposed benchmark may evaluate whether VLMs can recall publicly available historical knowledge, rather than requiring genuine chronological reasoning. 
If this were the case, then a simple retrieval-augmented pipeline should substantially improve performance by explicitly accessing external factual knowledge sources. 
Specifically, such a system could first extract salient visual anchors from the image (e.g., artifact patterns, clothing styles, architectural elements, or event-related objects), then retrieve candidate references from publicly accessible sources such as Wikimedia Commons or Wikipedia, and finally aggregate the retrieved metadata to infer the dynasty or year. Under the ``knowledge recall'' hypothesis, such retrieval augmentation would be expected to significantly reduce the difficulty of the Artifacts and News tasks.

\par To investigate this possibility, we implement a lightweight Retrieval-Augmented Generation (RAG) baseline on top of open-source VLMs. 
As shown in Algorithm ~\ref{alg:rag_pipeline}, the pipeline consists of three stages:
(1) extract visual descriptions from the image using the VLM itself,
(2) retrieve candidate evidence from public knowledge sources through keyword matching, and (3) feed the retrieved textual evidence back into the model to produce the final chronological prediction.

\par Table~\ref{tab:rag_results} summarizes the comparison between the original zero-shot setting and the retrieval-augmented setting. 
The results reveal that retrieval provides only limited and inconsistent improvements (e.g., Qwen$3$-VL-$8$B from $35.80$ to $36.96$ in News-Year Task). 
And it may degrade performance (e.g., Qwen$3$-VL-$8$B from $38.43$ to $32.81$, MiniCPM-V-$4.5$ from $44.93$ to $40.92$ in Artifacts–Chronological Localization Task). 
These findings suggest that our benchmark cannot be trivially solved through lightweight retrieval pipelines alone. 
While public factual knowledge may provide auxiliary support, the dominant challenge remains the ability to perform fine-grained chronological reasoning under ambiguous visual evidence. 
In particular, the persistent performance gap between retrieval-augmented open-source models and zero-shot closed models indicates that the benchmark evaluates more than simple memorization of public historical facts.

\begin{table}[t]
\centering
\footnotesize
\caption{Comparison between zero-shot (ZS) and retrieval-augmented settings.}
\label{tab:rag_results}
\begin{tabular}{lcccc}
\toprule
Model & Dynasty$_{ZS}$ & Dynasty$_{RAG}$ & Year$_{ZS}$ & Year$_{RAG}$ \\
\midrule
Qwen$3$-VL-$4$B & $40.14$ & $38.22$ & $32.86$ & $34.34$ \\
Qwen$3$-VL-$8$B & $38.43$ & $32.81$ & $35.80$ & $36.96$ \\
MiniCPM-V-$4.5$ & $44.93$ & $40.92$ & $31.53$ & $29.57$ \\
\bottomrule
\end{tabular}
\vspace{-1em}
\end{table}

\section{Conclusion}
\par In this paper, we introduced a comprehensive benchmark along with a newly constructed dataset specifically designed to evaluate the chronological reasoning capabilities of VLMs. 
Our work systematically investigates how models perform across three critical dimensions: Artifacts, Shortcuts, and News. 
Through extensive experiments on leading VLMs, we reveal that authentic chronological logic remains a significant hurdle for current multimodal systems.

\par Our findings yield three insights: \textbf{Persistence of Visual Shortcuts:} 
We demonstrate that models heavily rely on superficial cues, such as the ``grayscale equals old'' heuristic. When these shortcuts conflict with chronological facts, model reasoning often collapses, revealing a lack of deep historical understanding. 
\textbf{Domain-Specific Robustness:} 
Our dataset reveals that categories with distinct technological iteration features, such as electronics, provide stronger visual anchors that help models resist color bias. 
In contrast, more abstract domains remain highly susceptible to stylistic misleading. 
\textbf{Divergent Reasoning Bottlenecks:} 
We identify a performance disparity where models exhibit different limitations in cross-modal year alignment versus absolute chronological localization. 

\par Ultimately, our work provides a rigorous foundation for assessing chronological intelligence. 
These results highlight the urgent need for future training paradigms that decouple superficial image styles from intrinsic chronological semantics to achieve more robust multimodal reasoning.

\bibliographystyle{IEEEtran}
\bibliography{ref}

\end{document}